\pdfoutput=1
\documentclass[conference]{IEEEtran}
\usepackage[a4paper, total={184mm,239mm}]{geometry}
\usepackage{amsmath,amssymb,amsfonts,amsthm,enumitem}
\usepackage[ruled,linesnumbered]{algorithm2e}
\usepackage{xcolor}
\usepackage{algorithmic}
\usepackage{graphicx}
\usepackage{textcomp}
\usepackage{float}
\usepackage{placeins}
\usepackage{multirow}
\usepackage{dblfloatfix}
\usepackage{pifont}
\usepackage{colortbl}
\usepackage[normalem]{ulem}
\usepackage{booktabs}

\usepackage{adjustbox}
\usepackage{enumitem}
\usepackage[english]{babel}
\usepackage[utf8]{inputenc}
\usepackage[colorinlistoftodos]{todonotes}
\usepackage{balance}
\usepackage{comment}
\usepackage{pgfplots}
\usepackage{subcaption}
\usepackage{soul}
\algsetup{linenosize=\tiny}

\def\BibTeX{{\rm B\kern-.05em{\sc i\kern-.025em b}\kern-.08em
    T\kern-.1667em\lower.7ex\hbox{E}\kern-.125emX}}
    
\begin{document}

\title{Improving Reliability of Spiking Neural Networks through Fault Aware Threshold Voltage Optimization}
\author{\IEEEauthorblockN{Ayesha Siddique, Khaza Anuarul Hoque}
\IEEEauthorblockA{\textit{Department of Electrical Engineering and Computer Science}\\ \textit{University of Missouri,
Columbia, MO, USA}\\
ayesha.siddique@mail.missouri.edu, hoquek@missouri.edu}
}

\maketitle

\begin{abstract}
Spiking neural networks have made breakthroughs in computer vision by lending themselves to neuromorphic hardware. However, the neuromorphic hardware lacks parallelism and hence, limits the throughput and hardware acceleration of SNNs on edge devices. To address this problem, many systolic-array SNN accelerators (systolicSNNs) have been proposed recently, but their reliability is still a major concern. In this paper, we first extensively analyze the impact of permanent faults on the SystolicSNNs. Then, we present a novel fault mitigation method, i.e., fault-aware threshold voltage optimization in retraining (FalVolt). FalVolt optimizes the threshold voltage for each layer in retraining to achieve the classification accuracy close to the baseline in the presence of faults. To demonstrate the effectiveness of our proposed mitigation, we classify both static (i.e., MNIST) and neuromorphic datasets (i.e., N-MNIST and DVS Gesture) on a 256x256 systolicSNN with stuck-at faults. We empirically show that the classification accuracy of a systolicSNN drops significantly even at extremely low fault rates (as low as 0.012\%). Our proposed FalVolt mitigation method improves the performance of systolicSNNs by enabling them to operate at fault rates of up to 60\%, with a negligible drop in classification accuracy (as low as 0.1\%). Our results show that FalVolt is 2x faster compared to other state-of-the-art techniques common in artificial neural networks (ANNs), such as fault-aware pruning and retraining without threshold voltage optimization.
\end{abstract}


\section{Introduction}
\label{sec:introduction}
\vspace{1mm}
Spiking neural networks (SNNs) are a promising third generation of neural networks that ensure high algorithmic performance at low power. Their hardware acceleration require specialized architectures such as, SpiNNaker~\cite{painkras2013spinnaker}, and TrueNorth~\cite{akopyan2015truenorth}. However, these architectures lack parallelism in each core and efficient dataflows for maximizing the reuse of weight data. This limits their achievable throughput and robustness in resource-constrained devices (e.g., battery-driven autonomous cars). Towards this, leveraging SNNs on massively parallel hardware accelerators such as systolic arrays has proven to be an efficient solution~\cite{lee2020reconfigurable, guo2019systolic, chuang202090nm, tan2020power, leeparallel}. Systolic array SNN accelerators (systolicSNNs) are inspired by other state-of-the-art hardware accelerators~\cite{kung2019packing} which support fully parallel execution of artificial neural networks (ANNs). These accelerators have a $N$x$N$ dense grid of interconnected processing elements (PEs), which allows efficient parallel processing with the high spatio-temporal locality. Unlike ANNs, SNNs and their hardware accelerators are still in a relatively early phase of adoption~\cite{el2020securing} and thus ensuring the reliability of systolicSNNs is still considered a major research challenge. 





The systolicSNN hardware chips are manufactured using nanometer CMOS technologies~\cite{lee2021systolic}, which require a highly sophisticated manufacturing process. The imperfections in this process result in various manufacturing defects ranging from process variations to permanent faults such as stuck-at faults. The stuck-at faults affect the output of systolicSNNs in every execution cycle and hence, lead to significant accuracy loss as discussed in this paper.  Furthermore, the impact of large-scale failures such as dead synapse faults in SNNs has been thoroughly investigated~\cite{schuman2020resilience, vatajelu2019special}. However, analyzing such failures in the hardware require a fault model with higher abstraction to make the simulation traceable. Guo et al. investigated the fault resilience of SNNs trained with different coding schemes by using a synaptic stuck-at fault model~\cite{guo2021neural}. El-Sayed et al. analyzed the effect of these faults in a transistor-level design of leaky-integrate-and-fire (LIF) neuron~\cite{el2020spiking}. Other state-of-the-art works focus on bit flips in weight memories~\cite{spyrou2022reliability, putra2021respawn, vidya2022softsnn, venceslai2020neuroattack}. Conversely, the impact of stuck-at faults on systolicSNNs has not been investigated. 

The stuck-at faults are usually detected using post-fabrication testing for discarding the faulty manufactured chips. However, if a high number of manufactured chips are faulty, discarding them reduces the yield to a large extent. A potential solution is employing redundant executions (re-execution) to ensure correct outputs, but it leads to significant latency and energy overheads~\cite{vidya2022softsnn}. In the current resource-constrained nanoscale hardware paradigm, where the number of PEs has drastically increased to meet the robustness requirements of the end users, it is imperative to maximize the yield with an efficient and fault-tolerant systolicSNN. Recently, Mehul et al. proposed an astrocyte self-repair mechanism for stuck-at 0 weights in SNNs~\cite{rastogi2021self}. Other works are either focused on mitigating the transient faults in SNNs~\cite{putra2021respawn, rastogi2021self} or contemplated permanent fault mitigation in ANN accelerators~\cite{siddique2021exploring, zhang2018analyzing, abdullah2020salvagednn, kundu2020high}. However, a considerable research gap exists in mitigating the impact of permanent faults in systolicSNNs.

\indent \textit{\textbf{Novel contributions:}} In this paper, we present an extensive stuck-at fault vulnerability analysis and a novel fault mitigation method i.e., \emph{\ul{\textbf{fa}}u\ul{\textbf{l}}t-aware retraining through threshold \ul{\textbf{volt}}age optimization (\ul{\textbf{FalVolt}})}. FalVolt first sets the weights mapped to faulty PEs only as zero and then retrains weights mapped to non-faulty PEs while optimizing the threshold voltage for each layer to restore the classification accuracy close to its baseline. The optimized threshold voltage differs from the actual threshold voltage used in initial training. To demonstrate the effectiveness of our proposed FalVolt mitigation method, we used both static MNIST~\cite{cohen2017emnist}, and neuromorphic N-MNIST~\cite{orchard2015converting} and DVS128 Gesture~\cite{amir2017low} datasets. Our results show that FalVolt can operate at high fault rates of up to 60\% with a negligible impact on the classification accuracy compared to its baseline. We empirically show that FalVolt takes 2x fewer retraining epochs, and thus it is 2x faster in restoring the baseline accuracy compared to other state-of-art techniques such as fault-aware pruning and retraining. Note, fault-aware pruning and retraining and threshold voltage optimization have been conventionally used for ANN fault mitigation~\cite{zhang2018analyzing, abdullah2020salvagednn} and faster SNN convergence. However, to the best of our knowledge, this is the first work to employ fault-aware threshold voltage optimization for fault mitigation in SNNs.

The remainder of this paper is structured as follows: Section~\ref{sec:prelim} provides the preliminary information about SNNs and systolicSNNs. Section~\ref{sec:motivation} and Section~\ref{sec:techniques} present a motivational case study and the proposed FalVolt mitigation method for systolicSNNs, respectively. Section~\ref{sec:results} discusses the results for the fault vulnerability and mitigation. Finally, Section~\ref{sec:conclusion} concludes the paper.

\section{Background}
\label{sec:prelim}
\vspace{2mm}
This section provides a brief overview of the state-of-the-art SNNs and systolicSNNs for better understanding.

\vspace{0.05in}
\textbf{Spiking Neural Networks}: SNNs are bio-inspired artificial neural networks. Their working principle can be explained with a standard LIF model as follows: when the membrane potential $V_{t}$ of a presynaptic neuron exceeds a specific threshold voltage at time $t$, a post-synaptic spike is fired, and then, $V_{t}$ relaxes to the resting state ($V_{rest}$ $<$ threshold voltage) with a time constant $\tau$. $V_{t}$ maintains the resting state for a refractory time $t_{ref}$ before responding to the received spikes. The LIF-based SNNs learn the presynaptic weights but require manual tuning of the time constant in training. Furthermore, the time constant is typically chosen to be the same for all neurons, which limits the diversity of neurons and, thus, the expressiveness of the LIF-based SNNs. Recently, Fang et al. proposed to train the weights along with the time constant through an advanced LIF model, i.e., parametric leaky integrate-and-fire (PLIF)~\cite{fang2021incorporating}. Incorporating the learnable time constants through PLIF-based SNNs makes the network less sensitive to initial values and reduces the training time. 

\begin{figure}[!t]
	\centering
	\vspace{-0.01in}
	\includegraphics[width=0.9\linewidth]{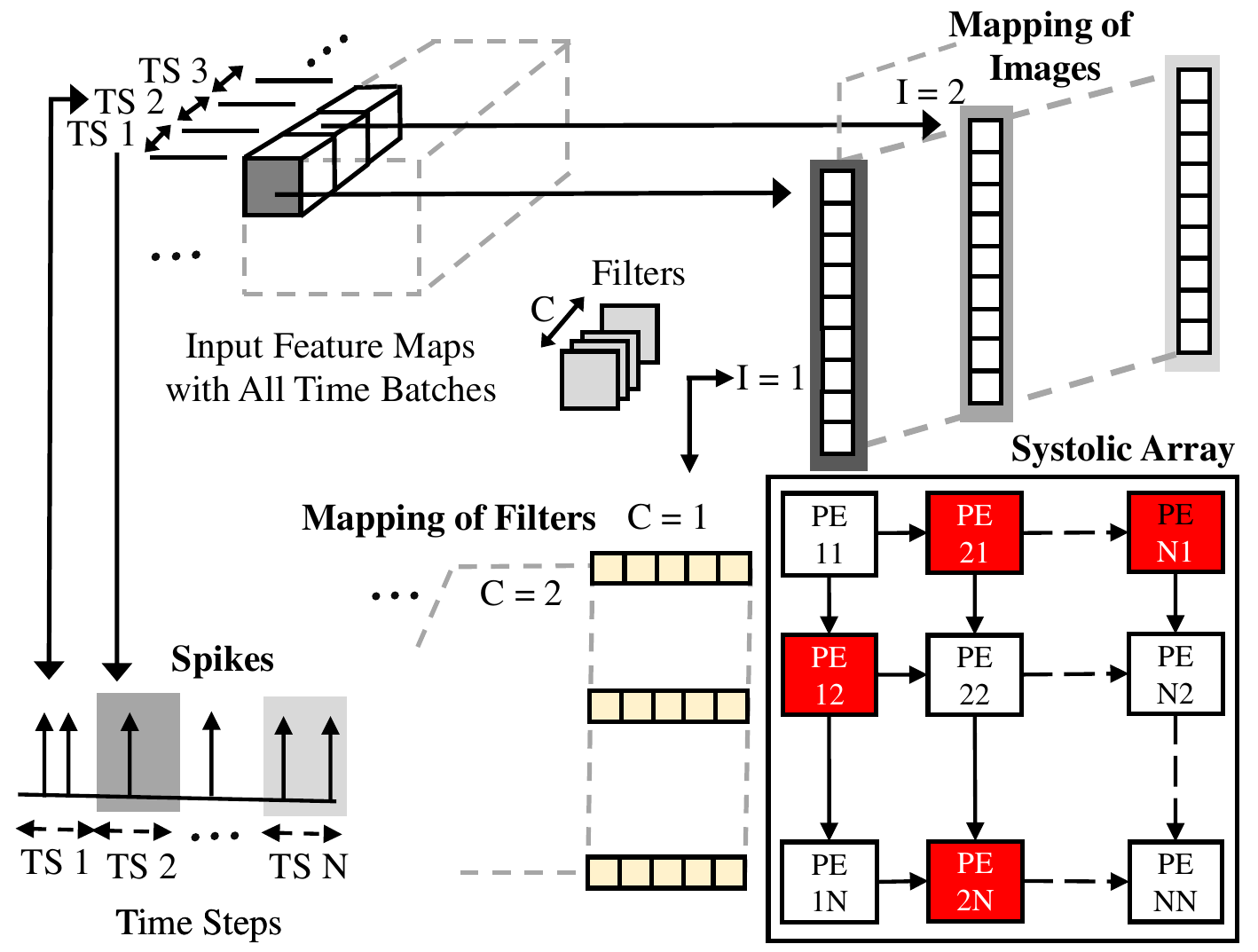}
	\caption{A systolicSNN with faulty processing elements (PEs) in red color and non-faulty PEs in white color}
	\label{fig:systolic_array_only}
	\vspace{-0.22in}
\end{figure}
\begin{figure}[!h]
	\centering
	\vspace{0.01in}
	\includegraphics[width=1\linewidth]{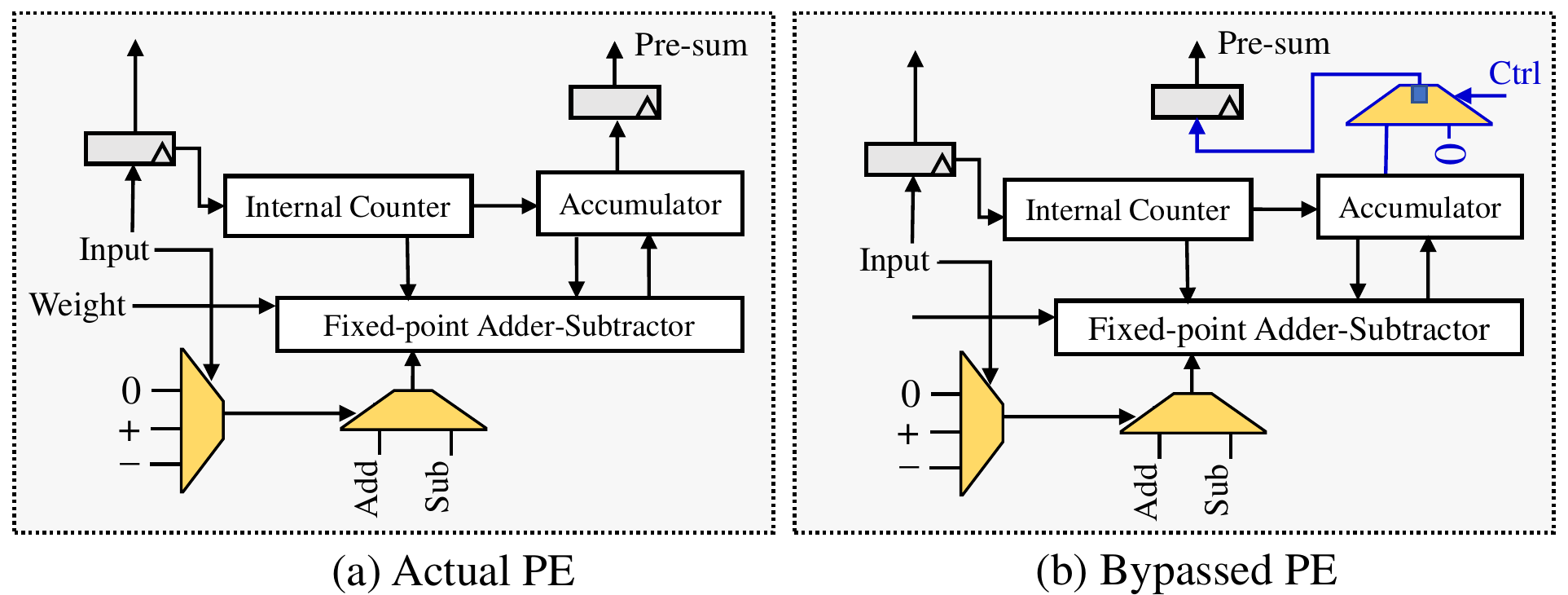}
	\caption{Processing element with actual and bypassed circuitry}
	\label{fig:PE}
	\vspace{-0.25in}
\end{figure}

\vspace{0.05in}
\textbf{Systolic-Array SNN Accelerators:} SystolicSNNs exploit the spatial and temporal parallelisms for which binary spike input, logical 1 or 0 propagate vertically across the systolic array. As shown in Fig.~\ref{fig:systolic_array_only}, the spike input is first divided into multiple time steps and then, all input values in a time step are mapped on one row of the systolic array. The input binary spikes pass through a dense $N$x$N$ grid of interconnected PEs in a clocked synchronized manner. The filter data is mapped and pre-stored in the PEs. Fig.~\ref{fig:PE}a shows the design of a standard PE in systolicSNNs. The PE accumulates 32-bit weight inputs under 1-bit binary spikes on an enable signal. The adder needed for the accumulation operation in systolicSNNs is cheaper than the multiplier needed for the multiplier-and-accumulator (MAC) unit in systolic-array ANN accelerators~\cite{guo2019systolic}~\cite{wang2020sies}. The lack of multipliers renders systolicSNNs energy efficient in comparison to systolic-array ANN accelerators. The PEs employs an addition and subtraction selection unit also for processing signed weights. Furthermore, an internal counter helps in counting the number of spikes in the inference phase. 

\section{Motivational Case Study} 
\label{sec:motivation}
\begin{figure}[!t]
     \centering
     \vspace{-0.01in}
     \begin{subfigure}[b]{0.23\textwidth}
         \centering
         \includegraphics[width=1\textwidth]{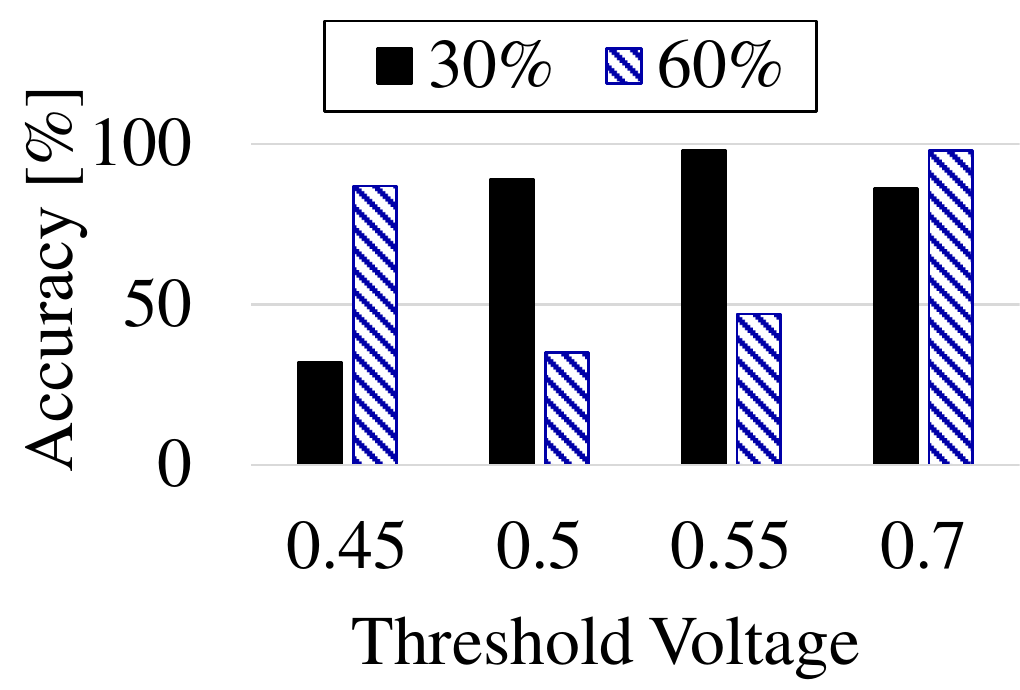}
         \caption{MNIST classification}
         \label{subfig:mitigation1a}
     \end{subfigure}
     \hfill
     \begin{subfigure}[b]{0.25\textwidth}
         \centering
         \includegraphics[width=0.9\textwidth]{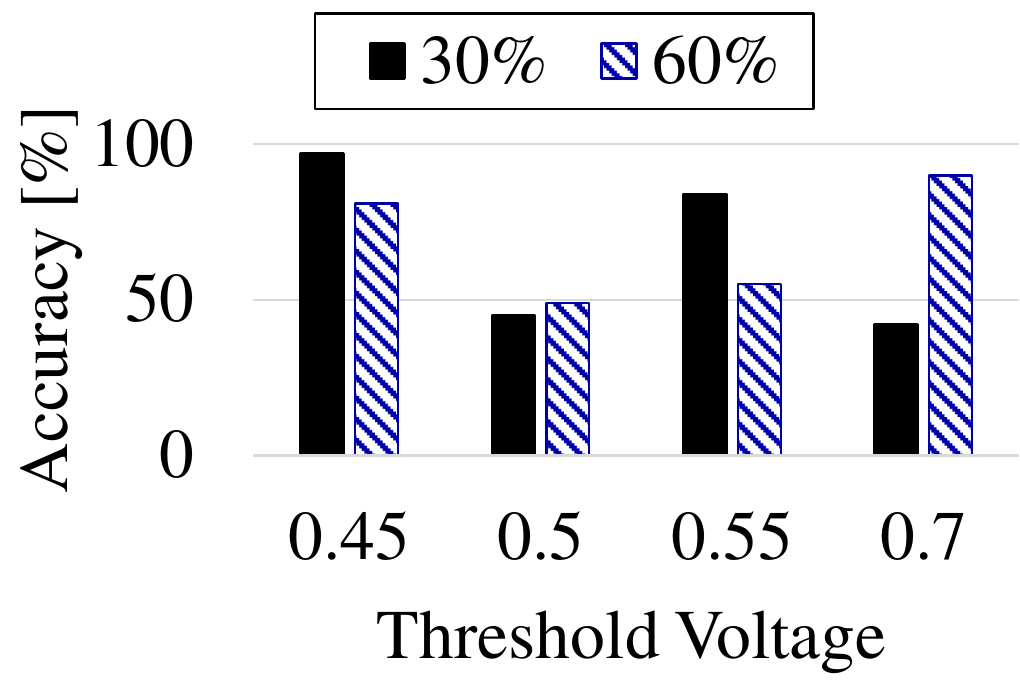}
         \caption{DVS128 Gesture classification}
         \label{subfig:mitigation1b}
     \end{subfigure}
        \vspace{-0.2in}
        \caption{Stuck-at fault mitigation using different threshold voltages ($V_{th}$), 30\% and 60\% of the total PEs are faulty in a 256x256 systolic-array SNN accelerator (systolicSNN)}
        \label{fig:mitigation1}
       \vspace{-0.1in}
\end{figure}

\vspace{-2mm}
To motivate the proposed FalVolt mitigation method, we begin by empirically analyzing the impact of different threshold voltages on the classification accuracy of a faulty systolicSNN. To do so, we first train a PLIF-SNN with the MNIST and DVS128 Gesture datasets. Then, we inject the stuck-at faults using different fault maps for 30\% and 60\% PEs in a 256x256 systolicSNN. Next, we run paralleled retraining simulations with different threshold voltages. As shown in Fig.~\ref{fig:mitigation1}, we observe that changing the threshold voltage from 1.0 to 0.55 and 0.7 values in retraining leads to 99\% classification accuracy with the MNIST dataset when even 30\% and 60\% PEs are faulty in a systolicSNN, respectively. However, retraining the same model with threshold voltage 0.45 and 0.5 leads to almost 73\% and 60\% accuracy loss when 30\% and 60\% PEs are faulty in a systolicSNN, respectively. In addition, 0.45 and 0.7 threshold voltages are most suitable for classifying the DVS128 Gesture dataset with a systolicSNN having 30\% and 60\% faults in PEs, respectively. However, retraining the same model with threshold voltages 0.7 and 0.5 leads to almost 60\% and 55\% accuracy loss when 30\% and 60\% PEs are faulty in a systolicSNN, respectively. Thus, selecting an appropriate threshold voltage for retraining the systolicSNN with high classification accuracy is imperative. Nevertheless, finding a suitable threshold voltage requires extensive retraining simulations, which may incur a significant amount of time. Motivated by this, we propose a novel fault-aware threshold voltage optimization technique in retraining for fault mitigation.

\section{Proposed Fault-Aware Threshold Voltage Optimization (FalVolt)}
\label{sec:techniques}

Our proposed FalVolt mitigation method improves the reliability of systolicSNNs by first setting the input pre-trained weights which map to the faulty PEs as zero. The fault locations are determined through post-fabrication tests on a systolicSNN chip. This initial step is similar to bypassing a PE using a multiplexer at the hardware level, as shown in Fig.~\ref{fig:PE}b, in systolicSNNs. With the bypass path enabled, the contribution of the faulty PEs to the column sum is skipped. However, bypassing single faulty PE may result in the pruning of multiple pre-trained weights due to the reuse of systolicSNNs in the data processing. Therefore, FalVolt next retrains the unpruned weights while optimizing the threshold voltage for each layer. 

The threshold voltage optimization saves the retraining time by eliminating the need for an exhaustive search for an appropriate threshold voltage. It makes SNN less sensitive to initial values and enhances and speeds up the learning. The optimized threshold voltage is used for all neurons in a layer to reduce the retrainable parameters and time. FalVolt optimizes the weights using the recursive gradient computations during both initial training and retraining. The weights mapped to faulty PEs are set as zero at the end of every retraining epoch. However, the threshold voltage is optimized for each layer during the retraining only, as discussed below:  

Lets consider $\textbf{r}$ as a ratio between the membrane potential $v$ and threshold voltage $\mathbb{\overline{V}}$. A neuron fires an output spike $\textbf{o}$ when $v$ exceeds $\mathbb{\overline{V}}$. Mathematically, this can be written as:
\vspace{-1mm}  
\begin{equation}
     \textbf{z}^t_l = {\textbf{r}^t_l} - 1 \ \  and\ \       \textbf{\textbf{o}}^t_l =  \begin{cases}
      1, & \text{if $\textbf{z}^t_l > 0.$}\\
      0, & \text{otherwise.}
    \end{cases}       
\end{equation}
\vspace{-1mm}  

\SetInd{0.5em}{0.5em}
\begin{algorithm}[!t] \small
\caption{FalVolt Mitigation Algorithm}
\label{alg:robust}
\DontPrintSemicolon
\algsetup{linenosize=\small}

\SetKwInOut{Input}{Inputs}\SetKwInOut{Output}{Outputs}
\Input{(i) pre-trained weights: wts; (ii) training data: trData; (iii) test data: tsData; (iv) fault maps: fmaps; (v) time steps: T; (vi) max retraining epochs: trEpochs; (vii) learning rate: $\eta$; }
\Output{Accuracy: acc;}
\begin{algorithmic}[1]
\STATE ind = FindPrunedWeightsIndices (fmaps, wts) \\
//Find indices of pruning weights from fault maps 

\STATE pWts = SetPrunedWeightsToZero(ind, wts) \\
//Assign zero to the pruning weights at above indices

\STATE (pVth, $\theta$) = parameterInnitialization() \\
//Initialize $\theta$ and threshold voltage parameters

\FOR{epochs = 0 : trEpochs - 1}{
    \FOR{t = 0 : T - 1}{
        \FOR{l = 0 : L - 1}{
            \STATE (nWts) = UpdateWeights (pWts, ts, trData) \\
            //Update weights with backpropagation
            \STATE (nVth) = UpdateVoltageThresh (pVth, ts, trData) \\
            //Update threshold voltage with backpropagation
        }\ENDFOR
    \STATE L = CalculateLoss(trData) \\
    // Calculate cross entropy loss 
    \STATE $\theta$ = $\theta$ - $\eta$ $\Delta$L \\
    //Update network parameter $\theta$
    }\ENDFOR
    \STATE nWts = SetUpdatedWeightsToZero(nWts, ind) \\
    //Assign zero to all pruning weights using indices in Step 1
}\ENDFOR

\STATE acc = CheckInferenceAccuracy(nWts, tsData) \\
//Check inference accuracy using new weights 

\RETURN (nWts, nVth, acc);
        
\end{algorithmic}
\end{algorithm}

Here, the notation $\textbf{x}^t_l$ represents the parameters of SNN in the l-th layer of the network at time step $t$. The discontinuous gradient $\frac{\partial \textbf{o}}{\partial \textbf{z}^t_l}$ is approximated with the surrogate function during error-backpropagation in retraining, similar to initial training. The term $\frac{\partial \textbf{o}}{\partial \textbf{z}^t_l}$ is expressed mathematically as:
\vspace{-2mm}
\begin{equation}
    \frac{\partial \textbf{o}^t_l}{\partial \textbf{z}^t_l} = \gamma 
    \max(0, 1 - |\textbf{z}^t_l|)
\end{equation}

\noindent where $\gamma$ is a constant denoting the maximum value of the surrogate function. During backpropagation, the threshold voltage $\mathbb{\overline{V}}$ is updated for layer $l$ as follows:
\vspace{-1mm}  
\begin{equation}
    \mathbb{\overline{V}}_l = \mathbb{\overline{V}}_{l - 1} \ - \eta \ \Delta \mathbb{\overline{V}}
\end{equation}
\vspace{-1mm}  
\noindent where $\eta$ represents the learning rate. Here, the gradient of threshold voltage $\Delta \mathbb{\overline{V}}$ for layer $l$ can be computed as: 
\vspace{-2mm}  

\begin{equation}
    \Delta \mathbb{\overline{V}}_l = \frac{\partial L}{\partial \mathbb{\overline{V}}_l} = \sum\limits_{t=0}^{T - 1} \frac{\partial L}{\partial \textbf{o}^t_l} \frac{\partial \textbf{o}}{\partial \textbf{z}^t_l} \frac{\partial \textbf{z}}{\partial \mathbb{\overline{V}}_l}=\sum\limits_{t=0}^{T - 1} \frac{\partial L}{\partial \textbf{o}^t_l} \frac{\partial \textbf{o}}{\partial \textbf{z}^t_l} (\frac{ - \mathbb{\overline{V}}_l \textbf{o}^{t - 1}_l - v^t_l}{\mathbb{\overline{V}}^2_l})
\end{equation}

\noindent where $L$ represents the cross entropy loss function defined by the mean square error. Algorithm~\ref{alg:robust} delineates the proposed FalVolt mitigation method. Lines 1-2 prunes the pre-trained weights mapped to the faulty PEs in systolicSNNs. Line 3 initializes the heavy step function $\theta$ and $\mathbb{\overline{V}}$. Lines 4-5 computes the un-pruned weights and $\mathbb{\overline{V}}$ with multiple epochs in back-propagation. The un-pruned weights and $\mathbb{\overline{V}}$ are optimized in each time-step for every layer in the PLIF-SNN, while calculating the gradient of loss function ($\Delta L$) in Line 10-11. Line 13 set the weights mapped to faulty PEs as zero at the end of each training epoch. It is interesting to note that setting the re-training epochs to zero makes the FalVolt equivalent to simple fault-aware pruning (FaP). FalVolt returns new optimized values for the unpruned weights (or the re-trained model), $\mathbb{\overline{V}}$ for each layer and the improved classification accuracy. Note, the proposed mitigation needs to be performed once only for the fabricated chip based on its unique fault map and thus, helps in avoiding the re-fabrication cost of the chips.



\section{Results and Discussions}
\label{sec:results}

This section discusses the results obtained from the fault vulnerability and mitigation analysis of systolicSNNs.

\subsection{Datasets and network architectures}
We adopted a static MNIST~\cite{cohen2017emnist}, and two neuromorphic N-MNIST~\cite{orchard2015converting} and DVS128 Gesture~\cite{amir2017low} datasets in this paper. Note that the SNN research community widely uses these datasets for evaluating the performance of SNNs~\cite{morris2022hyperspike,putra2021respawn}. As a classifier for N-MNIST and MNIST datasets, we use a PLIF-based SNN with two times repeated set of convolutional, batch normalization, spiking neurons, and pooling layers and also, two times a set of dropout, fully connected, and spiking neurons layers. The former set is repeated five times with the same architecture configuration in the classifier for the DVS128 Gesture dataset. Furthermore, an additional set of convolutional layer and spiking neurons layer is used for spike encoding the input images, inspired by~\cite{lee2020enabling}, in these architectures. We use the initialization parameters from~\cite{fang2021incorporating} to achieve the baseline accuracy i.e., 99\% for the MNIST~\cite{cohen2017emnist} and N-MNIST~\cite{orchard2015converting} datasets, and 97\% for DVS128 Gesture~\cite{amir2017low} dataset, prior to fault injection in the inference phase. For systolicSNN inference, we developed a 256x256 grid of PEs in VHDL with bypass circuitry that incurs only 8\% area overhead.

\subsection{Simulation Methodology}
\begin{figure}[!t]
	\centering
	\includegraphics[width=\columnwidth]{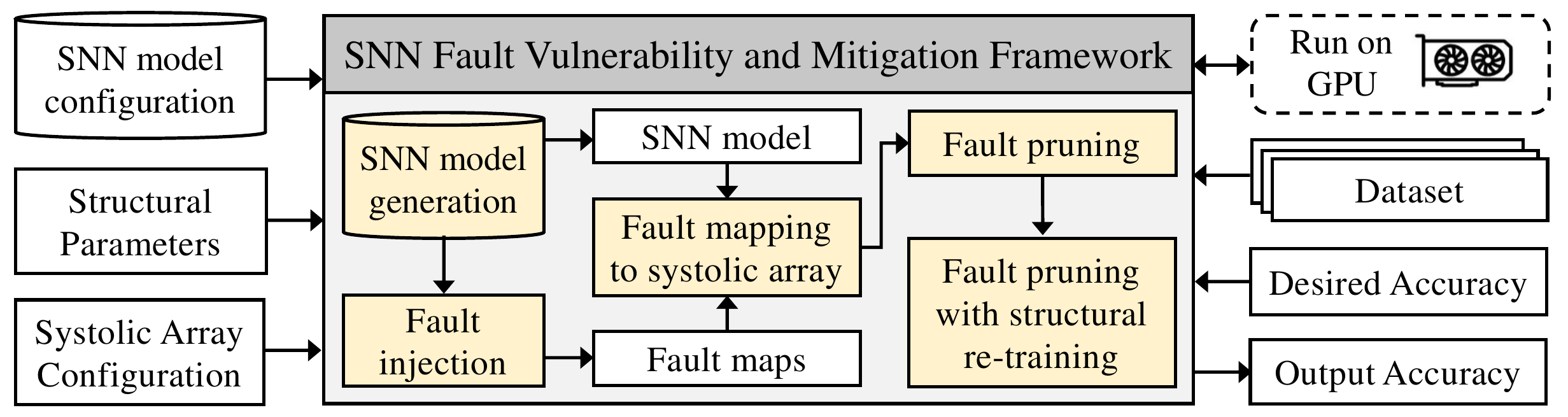}
	\caption{Experimental setup and tool flow} 
	\label{fig:toolflow}
	\vspace{-0.25in}
\end{figure}

Fig.\ref{fig:toolflow} illustrates the tool-flow used for fault vulnerability and mitigation analysis in this paper. First, the SNN models are trained with their baseline accuracies. Next, the stuck-at faults are injected into the accumulator outputs of PEs using different fault maps. Then, the fault pruning is applied by setting the weights mapped to the faulty PEs as zero. Finally, fault mitigation through re-training with layer-wise threshold voltage optimization is employed using Algorithm~\ref{alg:robust}. All simulations are conducted using NVIDIA GeForce RTX 2080 Ti GPU on Intel Core i9-10900kF operating at 3.06 GHz with 32 GB RAM.

\begin{figure*}[!t]
     \centering
     \vspace{-0.01in}
     \begin{subfigure}[b]{0.33\textwidth}
         \centering
         \includegraphics[width=1\textwidth]{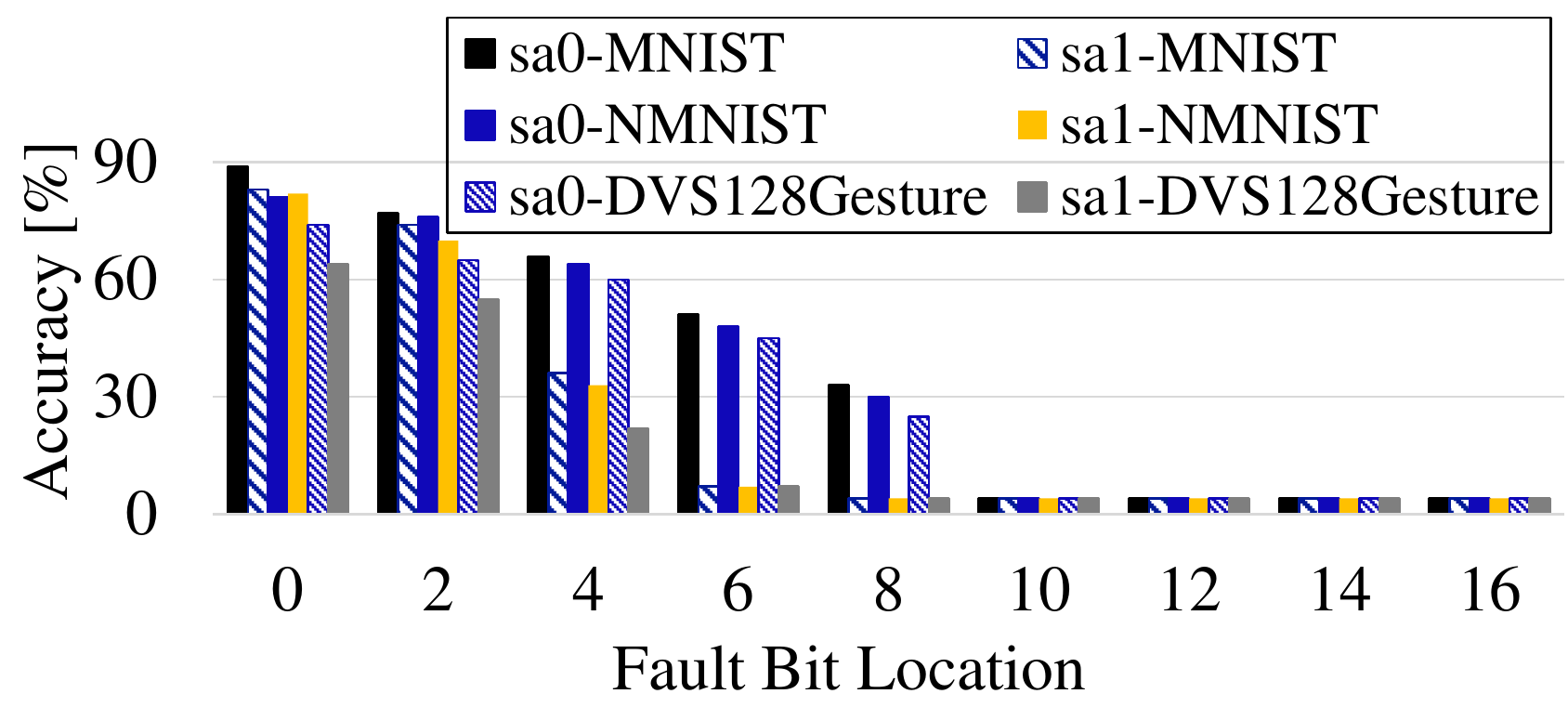}
         \caption{Accuracy vs Fault Bit Locations}
         \label{subfig:stuck}
     \end{subfigure}
     \hfill
     \begin{subfigure}[b]{0.35\textwidth}
         \centering
         \includegraphics[width=1\textwidth]{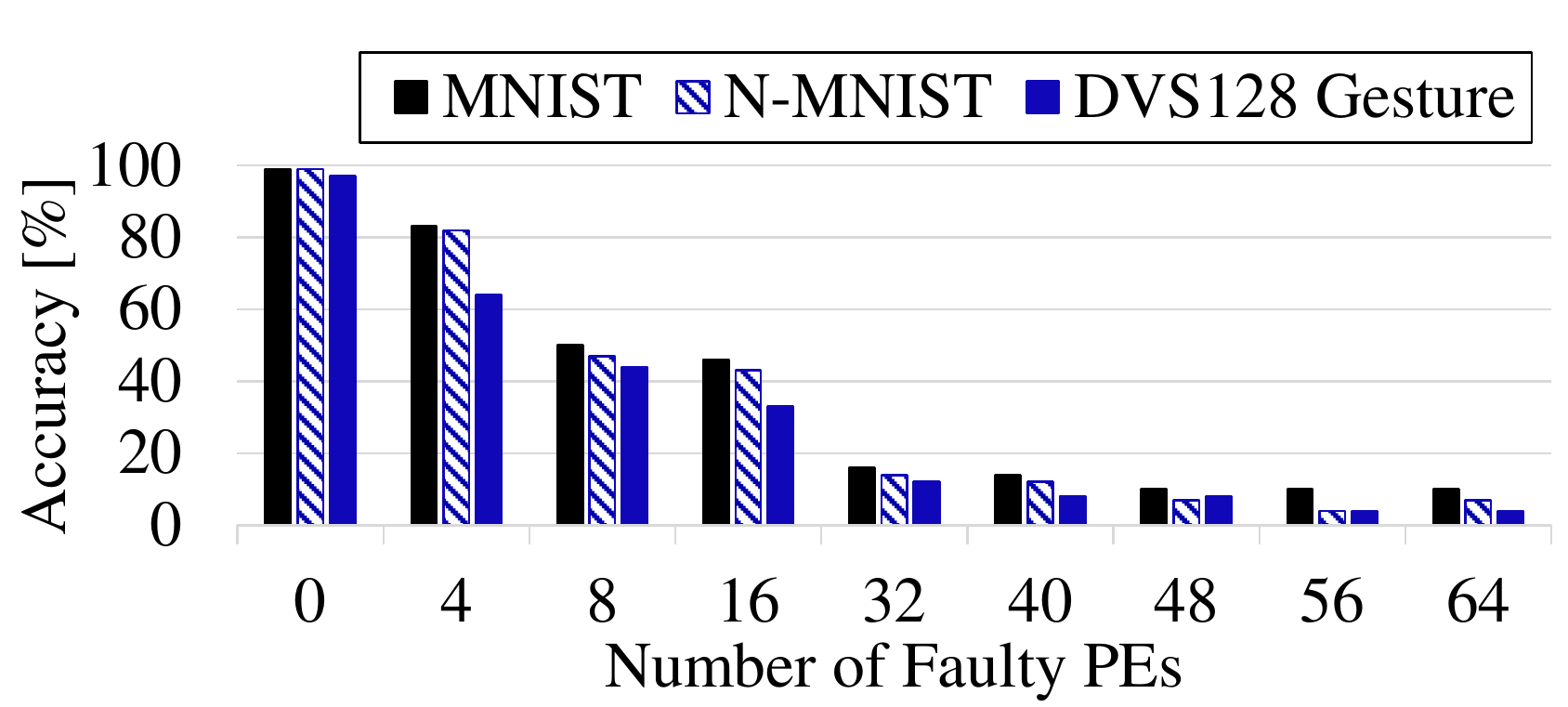}
         \caption{Accuracy vs number of faulty PEs}
         \label{subfig:res1a}
     \end{subfigure}
     \hfill
     \begin{subfigure}[b]{0.3\textwidth}
         \centering
         \includegraphics[width=1\textwidth]{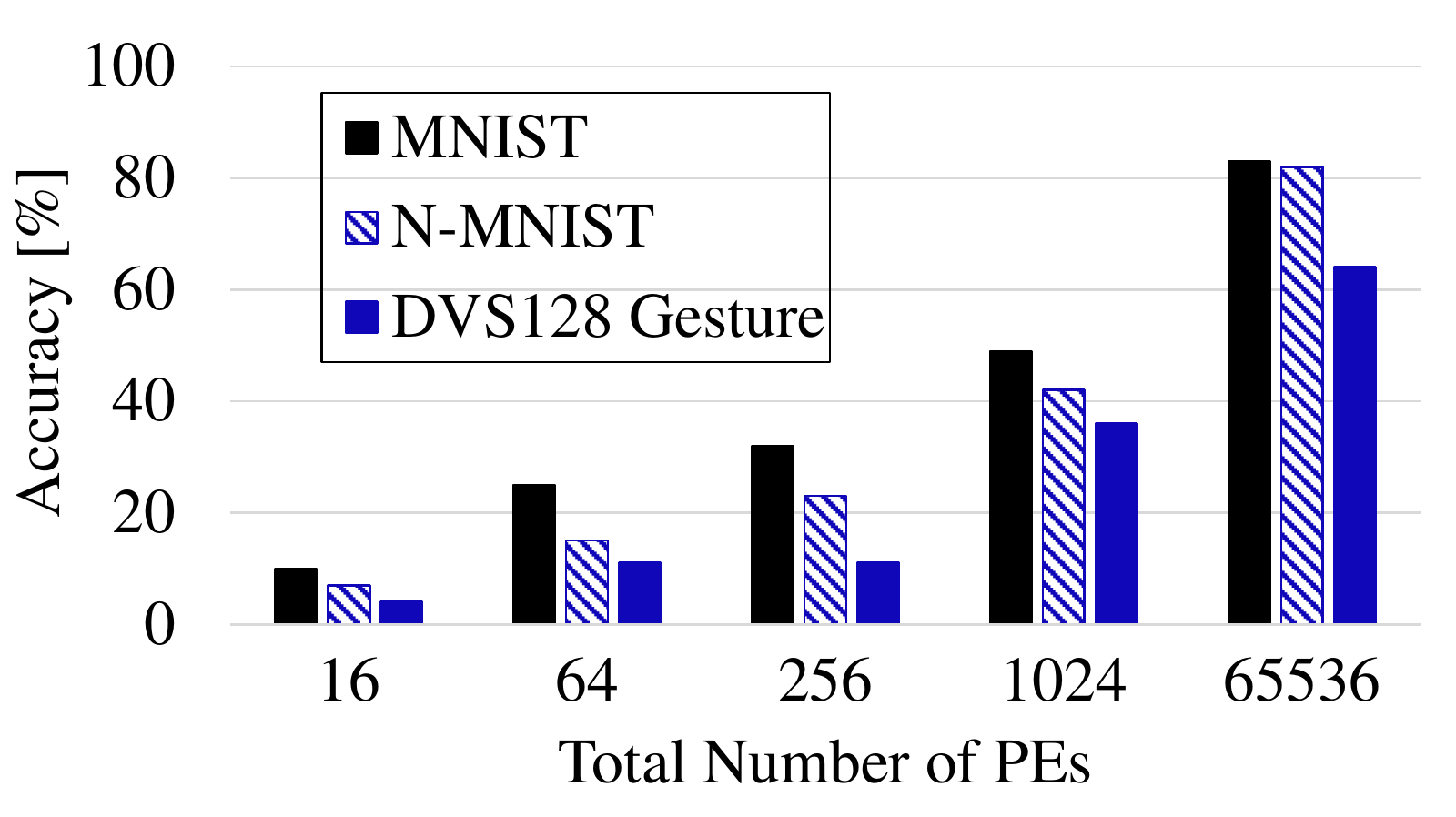}
         \caption{Accuracy vs size of systolic array}
         \label{subfig:res1b}
     \end{subfigure}     
        \vspace{-0.05in}
        \caption{Stuck-at fault vulnerability analysis of a 256x256 systolic-array based SNN accelerator (systolicSNN).}
        \label{fig:res1}
        \vspace{-0.1in}
\end{figure*}    
    
\begin{figure*}[!t]
     \centering
     \vspace{-0.05in}
     \begin{adjustbox}{minipage=\linewidth,width=18cm, height=2.5cm}
     \begin{subfigure}[b]{0.322\textwidth}
         \centering
         \includegraphics[width=1\textwidth]{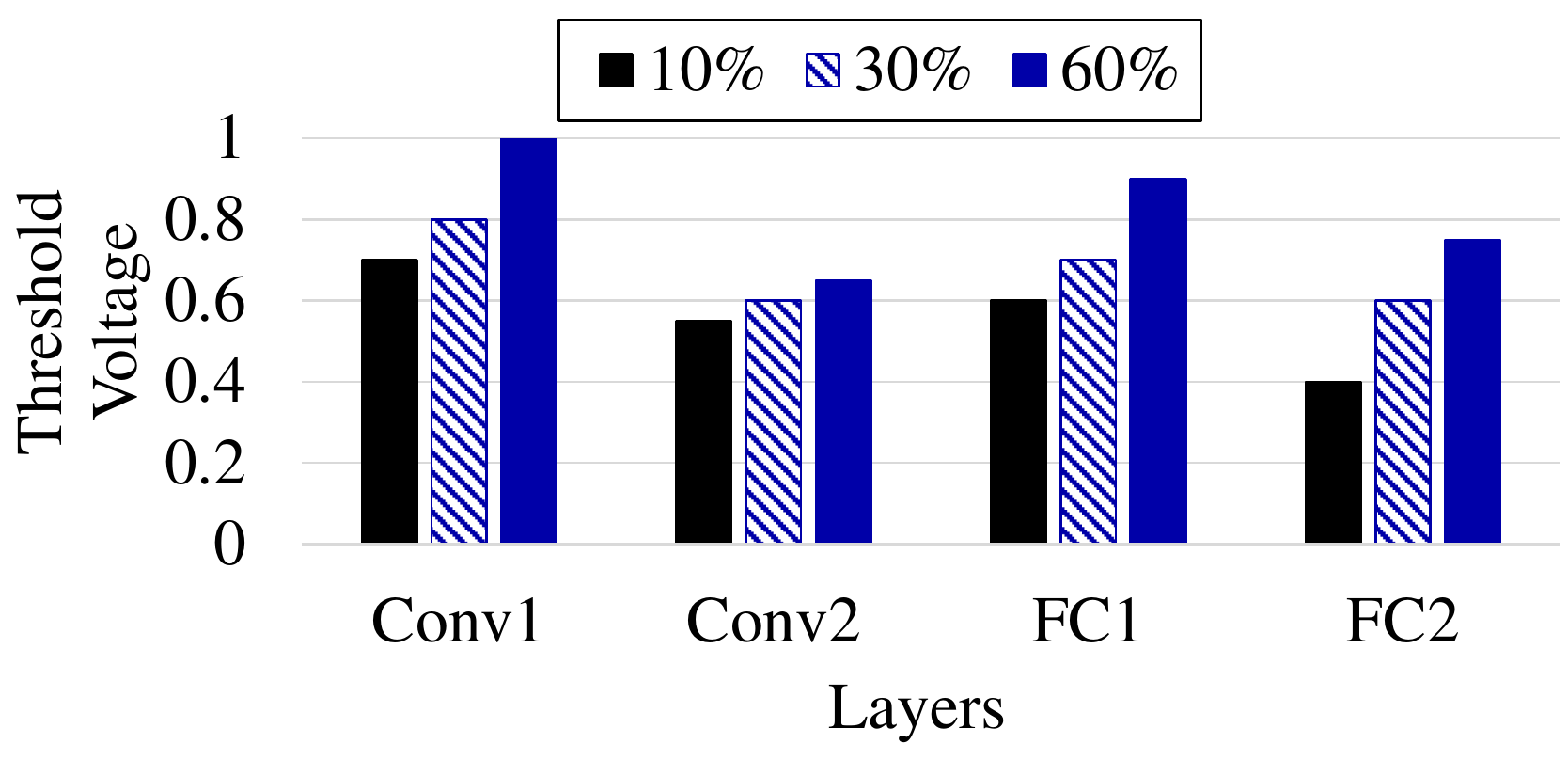}
         \caption{MNIST~\cite{cohen2017emnist} classification}
         \label{subfig:mitigation1a}
     \end{subfigure}
     \hfill
     \begin{subfigure}[b]{0.33\textwidth}
         \centering
         \includegraphics[width=1\textwidth]{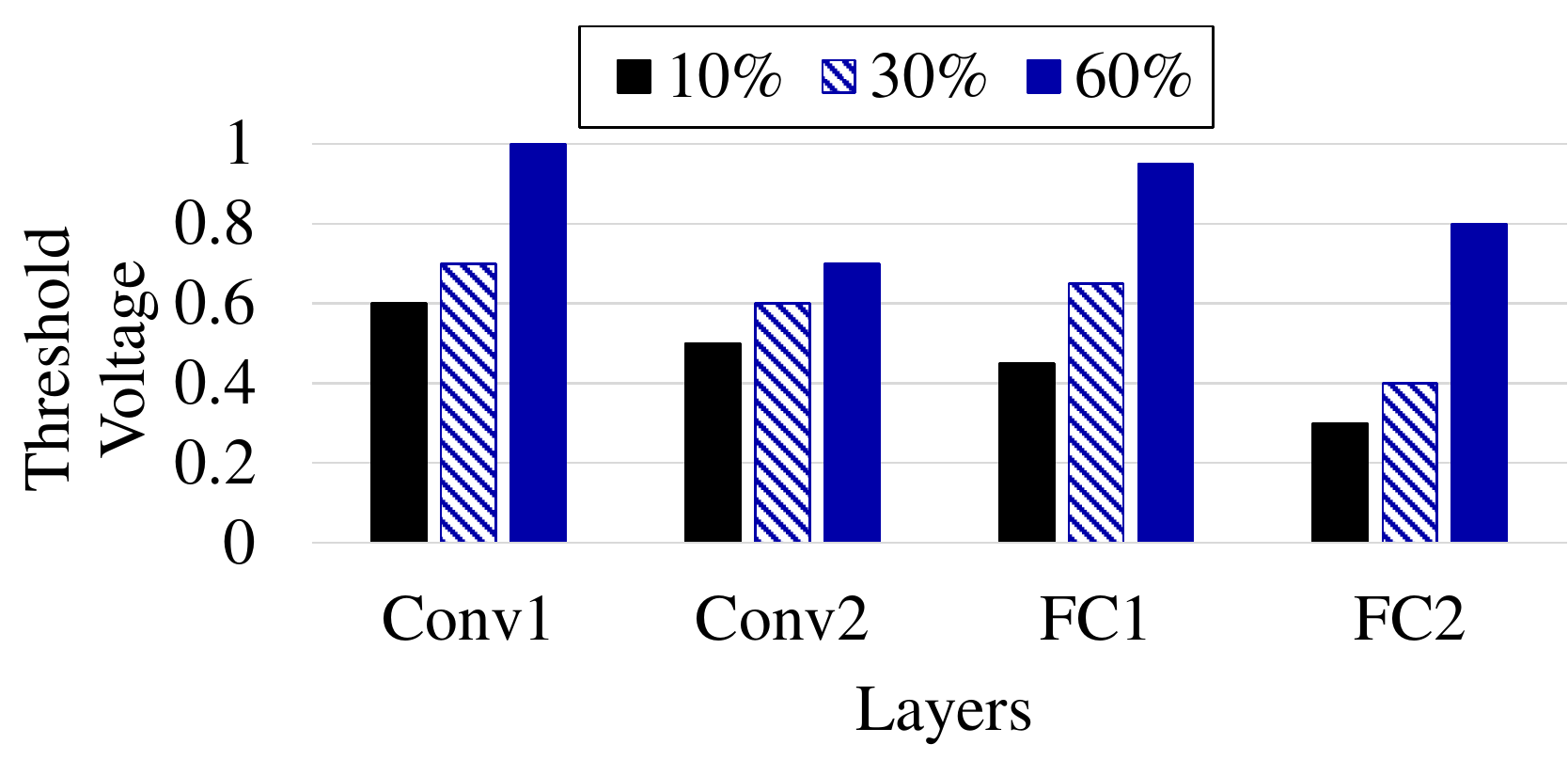}
         \caption{N-MNIST~\cite{orchard2015converting} classification}
         \label{subfig:mitigation1b}
     \end{subfigure}     
     \hfill
     \begin{subfigure}[b]{0.33\textwidth}
         \centering
         \includegraphics[width=1\textwidth]{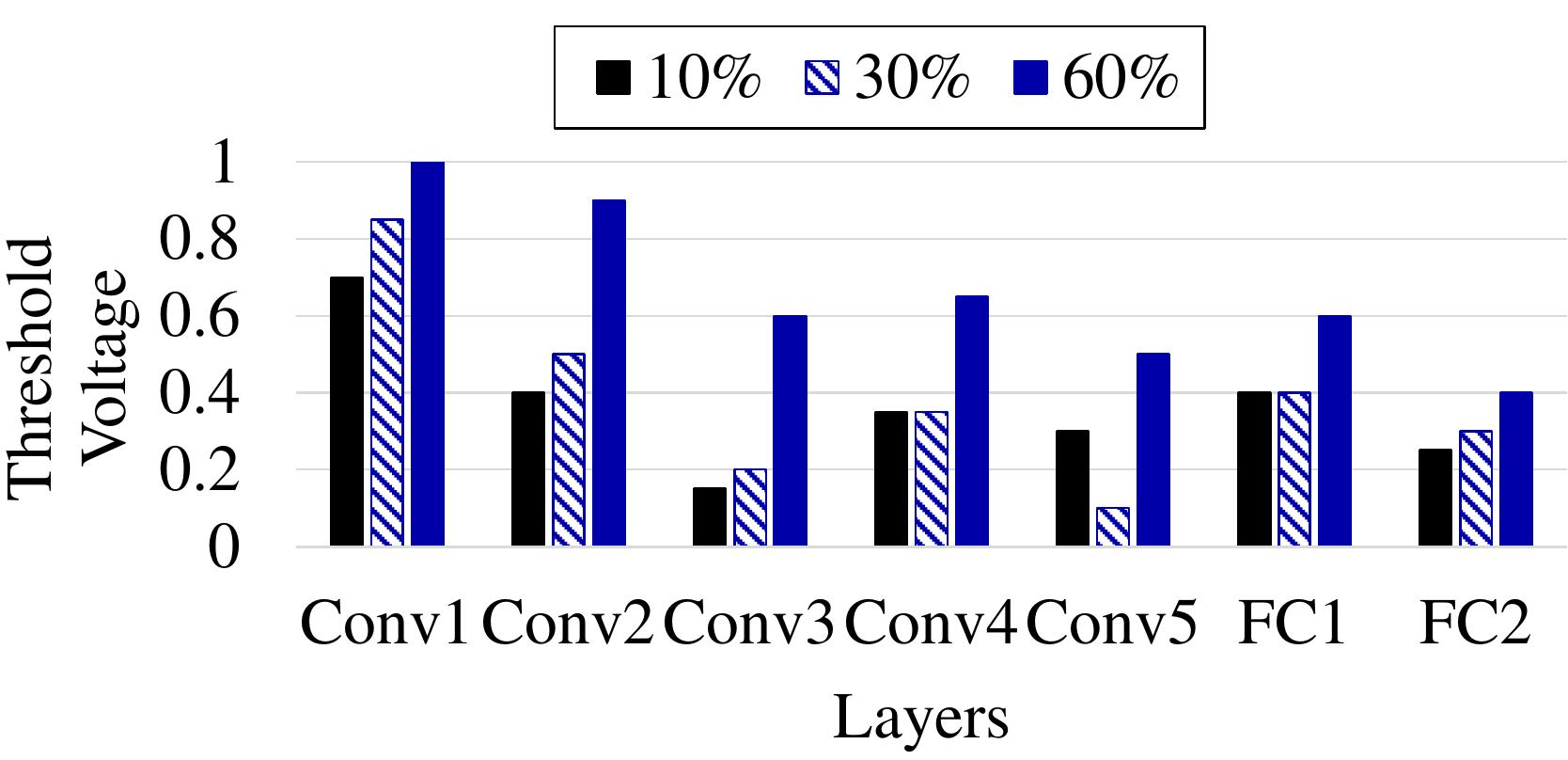}
         \caption{DVS128 Gesture~\cite{amir2017low} classification}
         \label{subfig:mitigation1b}
     \end{subfigure}
        \vspace{-0.04in}
        \caption{Optimized threshold voltage for hidden convolutional and fully connected layers using FalVolt, when 0\%, 10\%, 30\% and 60\% of the total PEs are faulty in a 256x256 systolic-array SNN accelerator (systolicSNN)}
        \label{fig:mitconv}
        \end{adjustbox}
        \vspace{-0.17in}
\end{figure*}


\subsection{Fault vulnerability analysis}
\label{subsubsec:vulnerability}
To investigate the stuck-at faults vulnerability in systolicSNNs, we extensively analyze their impact by varying the location of fault bits, the number of faulty PEs, and the size of the systolic array as follows.

\uline{Varying location of fault bits:} 
Before running extensive simulations for fault mitigation, we first identify the most vulnerable bits to the stuck-at faults in the PEs of a 256x256 systolicSNN. For this purpose, we generate the fault maps such that the stuck-at 0 and stuck-at 1 faults are injected in different output bit positions of the accumulator inside the PEs. Note, fault injection with fault maps is a common practice for analyzing the fault vulnerabilities in systolic arrays~\cite{hanif2020dependable}. Fault maps can be generated using post-fabrication testing in a real-world scenario. It is worth mentioning that we inject faults in the output of the accumulator, which is the main arithmetic component of the PEs. As shown in Fig.~\ref{subfig:stuck}, our analysis reveals that stuck-at faults in most significant bits (MSBs) affect the classification accuracy more than the stuck-at faults in the least significant bits (LSBs). The reason is that the systolic array is reused for different layers; therefore, a single unmasked fault in a PE of a particular layer affects all the connected nodes in the subsequent layers, decreasing the overall classification accuracy. We also observe that a stuck-at 1 fault in MSB causes almost 80\% accuracy loss, which is higher than the same fault in LSB when classifying the MNIST, N-MNIST, and DVS128 Gesture datasets. It is worth noticing that stuck-at 1 faults are more perturbing than stuck-at 0 faults in systolicSNN, similar to systolic array ANN accelerators~\cite{siddique2021exploring}.

\uline{Varying number of faulty PEs:} Next, we perform the fault simulations by considering a random distribution of the stuck-at faults across a 256x256 systolicSNN. We vary the fault rates by varying the number of faulty PEs in each experiment and running each experiment 8 times. The number of faulty PEs stays the same for all iterations in an experiment. Furthermore, each iteration uses a distinct fault map. In the following section, the faults are injected in the higher-order bits (i.e., MSBs) of the accumulator outputs in PEs to perform the worst-case analysis. Moreover, the average classification accuracies for all iterations in an experiment are recorded. As shown in Fig.~\ref{subfig:res1a}, our results demonstrate that \textit{even 8 faulty PEs (i.e., 0.012\% of total PEs)} can lead to an accuracy drop from 99\% to 50\%, 99\% to 47 \% and 97\% to 44\% in the MNIST, N-MNIST and DVS128 Gesture classification, respectively. Hence, the classification of both static and neuromorphic datasets is prone to stuck-at faults.

\uline{Varying size of the systolic array:} For further extensive fault vulnerability study, we analyze the impact of stuck-at faults across different sizes of \textit{N}x\textit{N} systolic arrays i.e., 4x4, 8x8, 16x16, 32x32 and 64x64. As shown in Fig.~\ref{subfig:res1b}, our analysis reveals that stuck-at faults in a small-sized systolic array cause more accuracy loss as compared to a large-sized systolic array. For example, 4 faulty PEs units in an 8x8 systolic array (having 16 PEs) lead to 89\%, 92\% and 93\% accuracy loss in the MNIST, N-MNIST and DVS128 Gesture classification, respectively. However, SNN classification with a 256x256 systolic array, having the same fault configuration, results in almost 16\%, 17\%, and 33\% accuracy loss only. This is due to the fact that decreasing the size of the systolic array increases its chances for re-usability and hence, the reoccurrence of the permanent faults in every execution cycle. 




Our analysis shows that DVS128 Gesture is more vulnerable to faults when compa
red to the MNIST and N-MNIST datasets, even though their baseline accuracies are the same. As shown in Fig.~\ref{subfig:res1a}, the classification accuracy of DVS128 Gesture remains comparatively lower than other datasets in the presence of stuck-at faults. Also, the accuracy loss associated with the DVS128 Gesture dataset is comparatively higher than other datasets in Fig.~\ref{subfig:res1b}. However, a higher number of stuck-at faults can render performance penalties unacceptable in all cases. 

\vspace{-0.05in}
\subsection{Fault mitigation analysis}
In this section, we study the performance of FalVolt and compare it with the state-of-the-art techniques common for ANNs. Specifically, we compare FalVolt with fault-aware pruning (FAP) and fault-aware pruning with retraining without threshold voltage optimization (FaPIT).  


\begin{figure*}[!t]
     \centering
     \vspace{-0.01in}
    \begin{adjustbox}{minipage=\linewidth,width=18cm, height=2.1cm}
     \begin{subfigure}[b]{0.322\textwidth}
         \centering
         \includegraphics[width=1\textwidth]{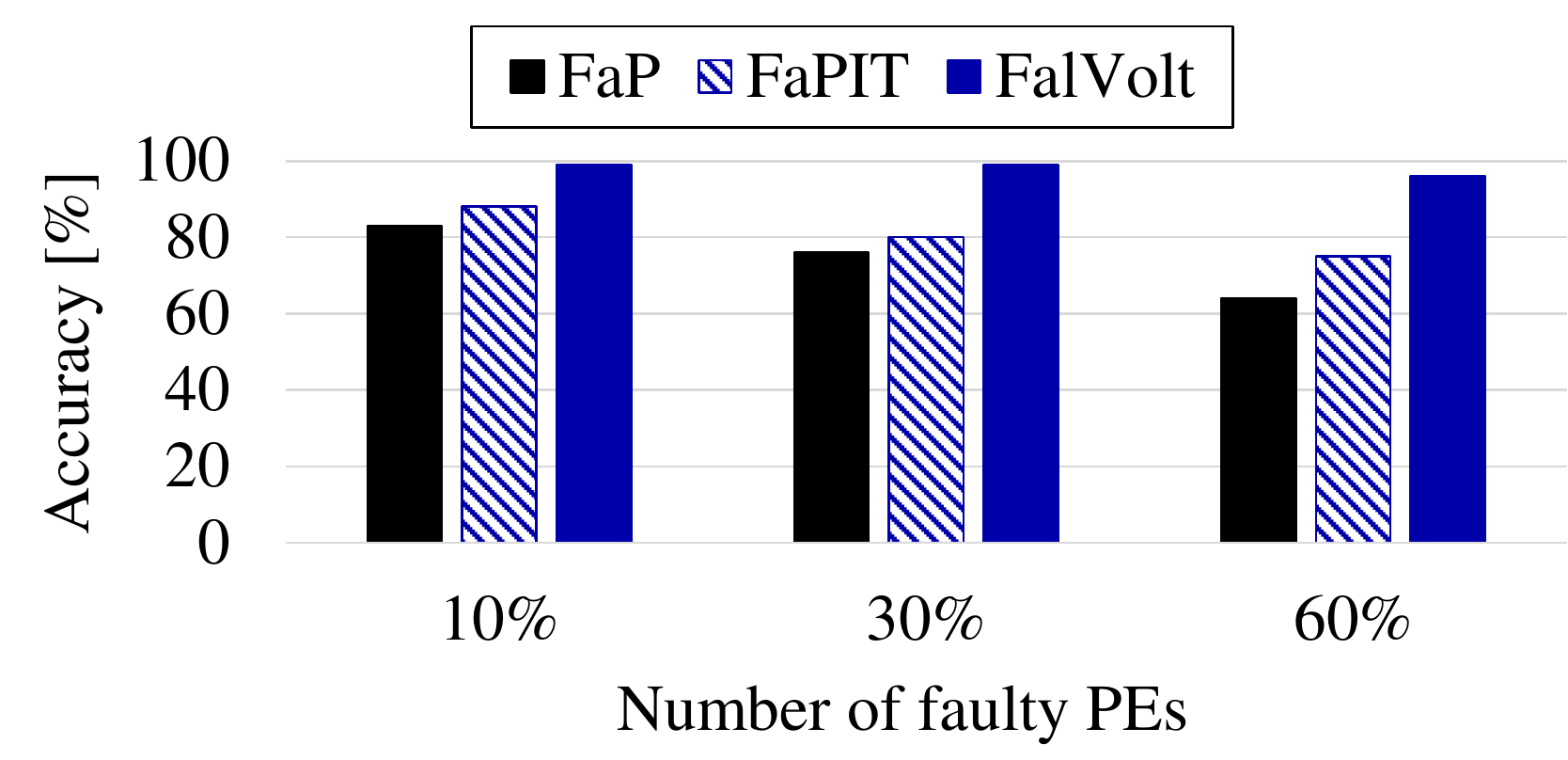}
         \caption{MNIST~\cite{cohen2017emnist} classification}
         \label{subfig:mitigation2a}
     \end{subfigure}
     \hfill
     \begin{subfigure}[b]{0.33\textwidth}
         \centering
         \includegraphics[width=1\textwidth]{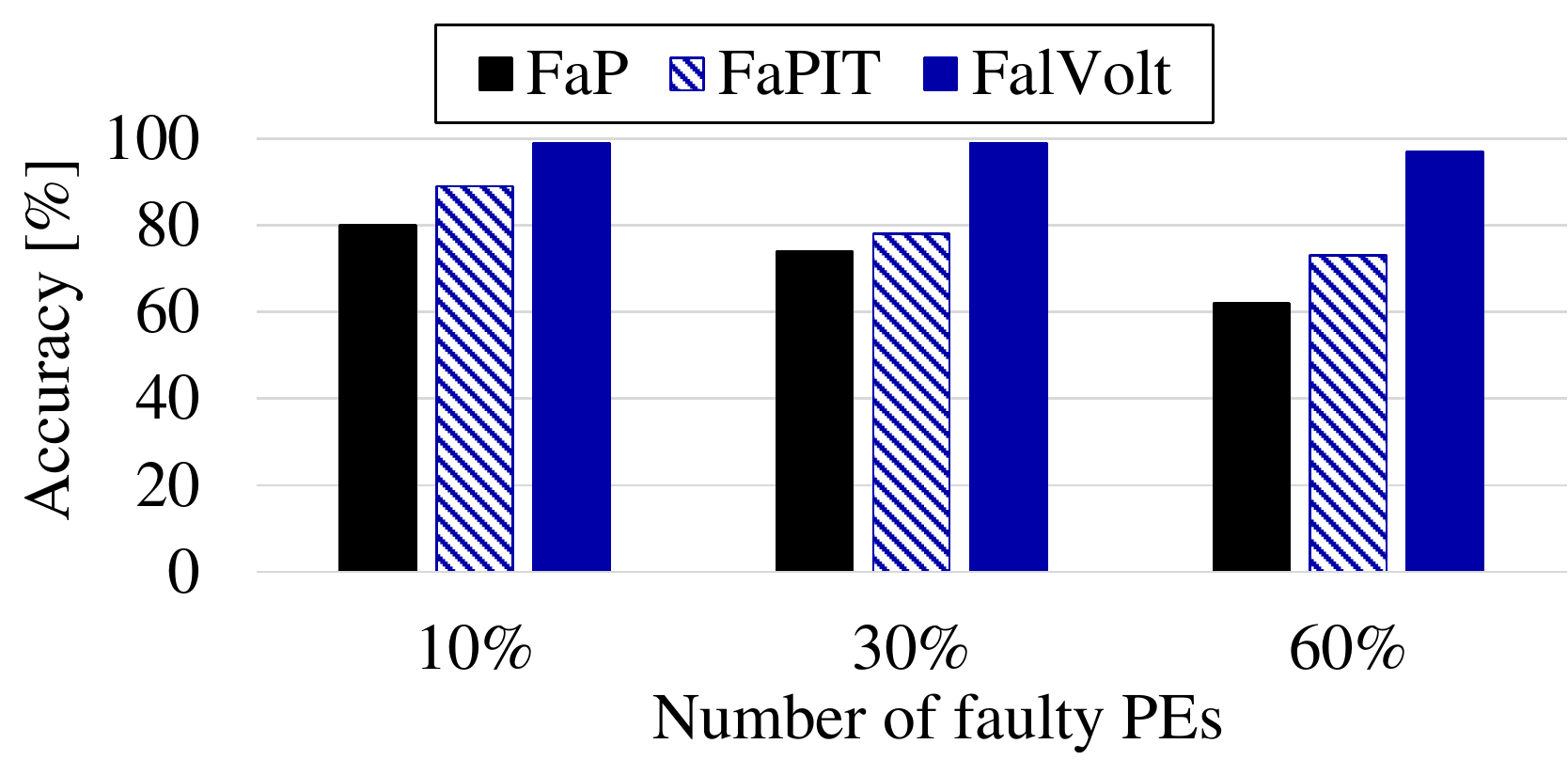}
         \caption{N-MNIST~\cite{orchard2015converting} classification}
         \label{subfig:mitigation2b}
     \end{subfigure}     
     \hfill
     \begin{subfigure}[b]{0.33\textwidth}
         \centering
         \includegraphics[width=1\textwidth]{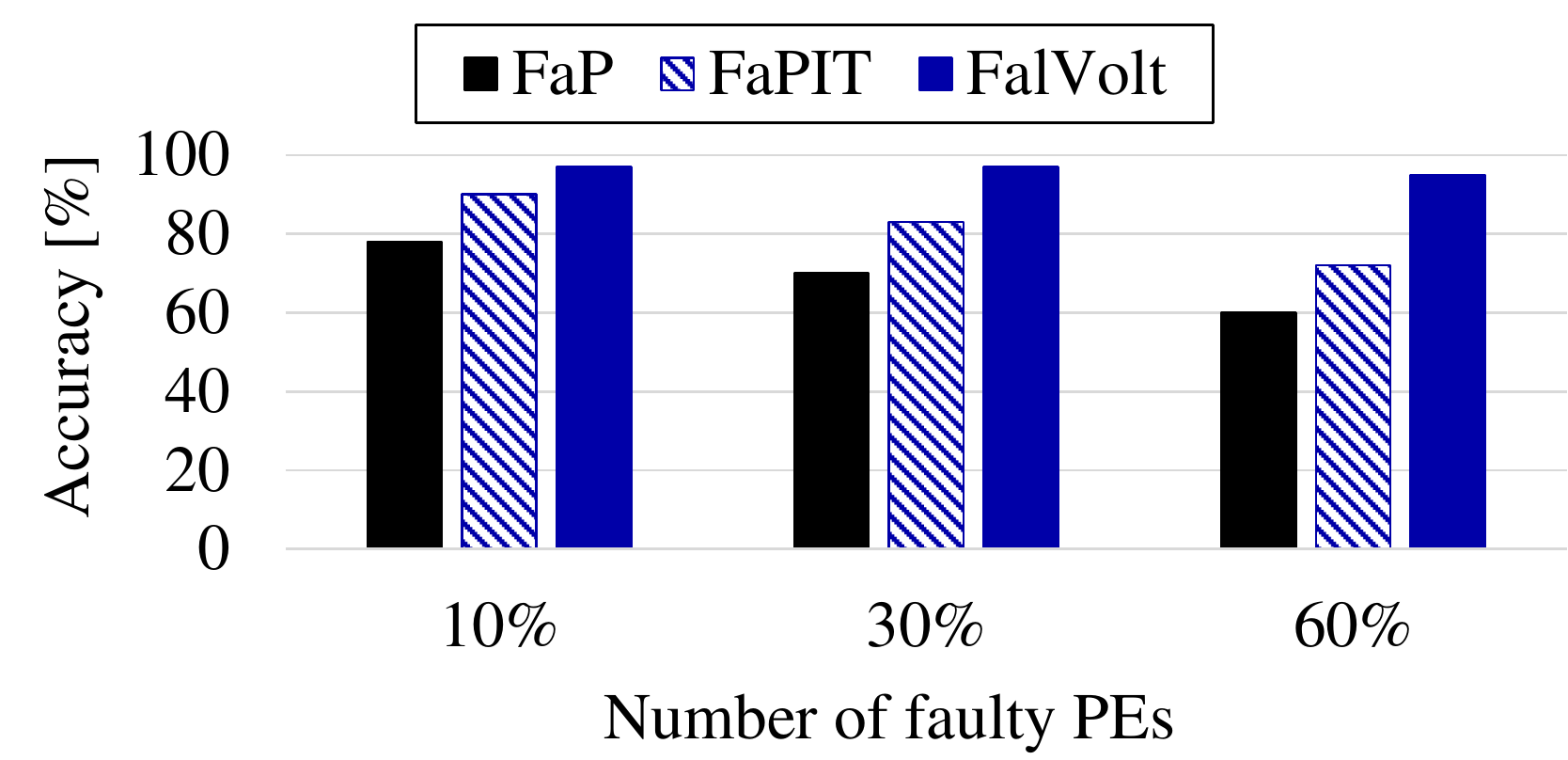}
         \caption{DVS128 Gesture~\cite{amir2017low} classification}
         \label{subfig:mitigation2c}
     \end{subfigure}
        \vspace{-0.05in}
        \caption{Stuck-at fault mitigation using FaP, FaPIT (using threshold voltage as 1.0) and FalVolt, when 0\%, 10\%, 30\% and 60\% of the total PEs are faulty in a 256x256 systolic-array SNN accelerator (systolicSNN)}
        \label{fig:mitcomp}
        \end{adjustbox}
        \vspace{-0.1in}
\end{figure*}    

\begin{figure*}[!t]
     \centering
     \vspace{-0.01in}
     \begin{adjustbox}{minipage=\linewidth,width=18cm, height=2cm}
     \begin{subfigure}[b]{0.322\textwidth}
         \centering
         \includegraphics[width=1\textwidth]{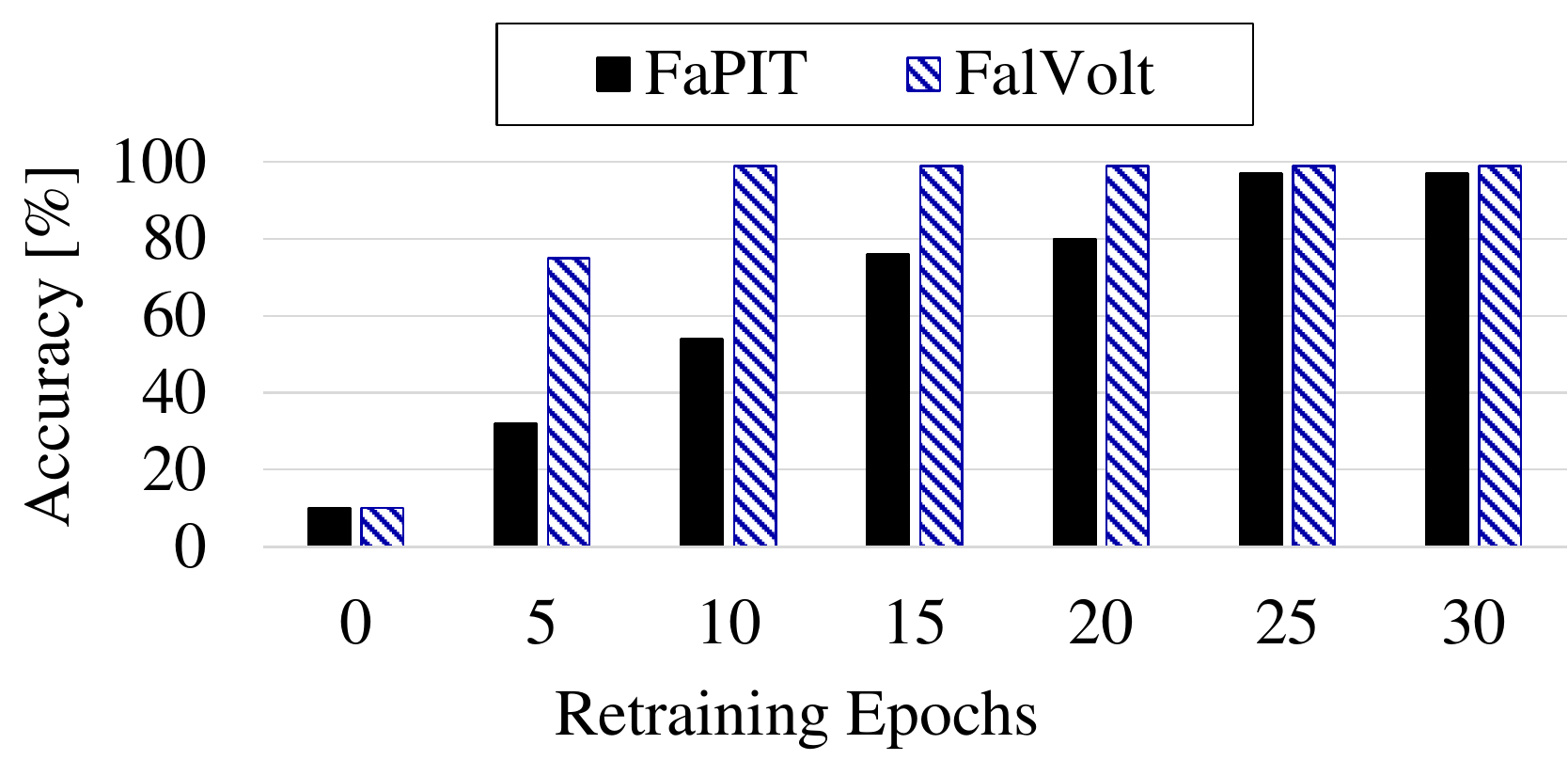}
         \caption{MNIST~\cite{cohen2017emnist} classification}
         \label{subfig:mitepoch11}
     \end{subfigure}
     \hfill
     \begin{subfigure}[b]{0.33\textwidth}
         \centering
         \includegraphics[width=1\textwidth]{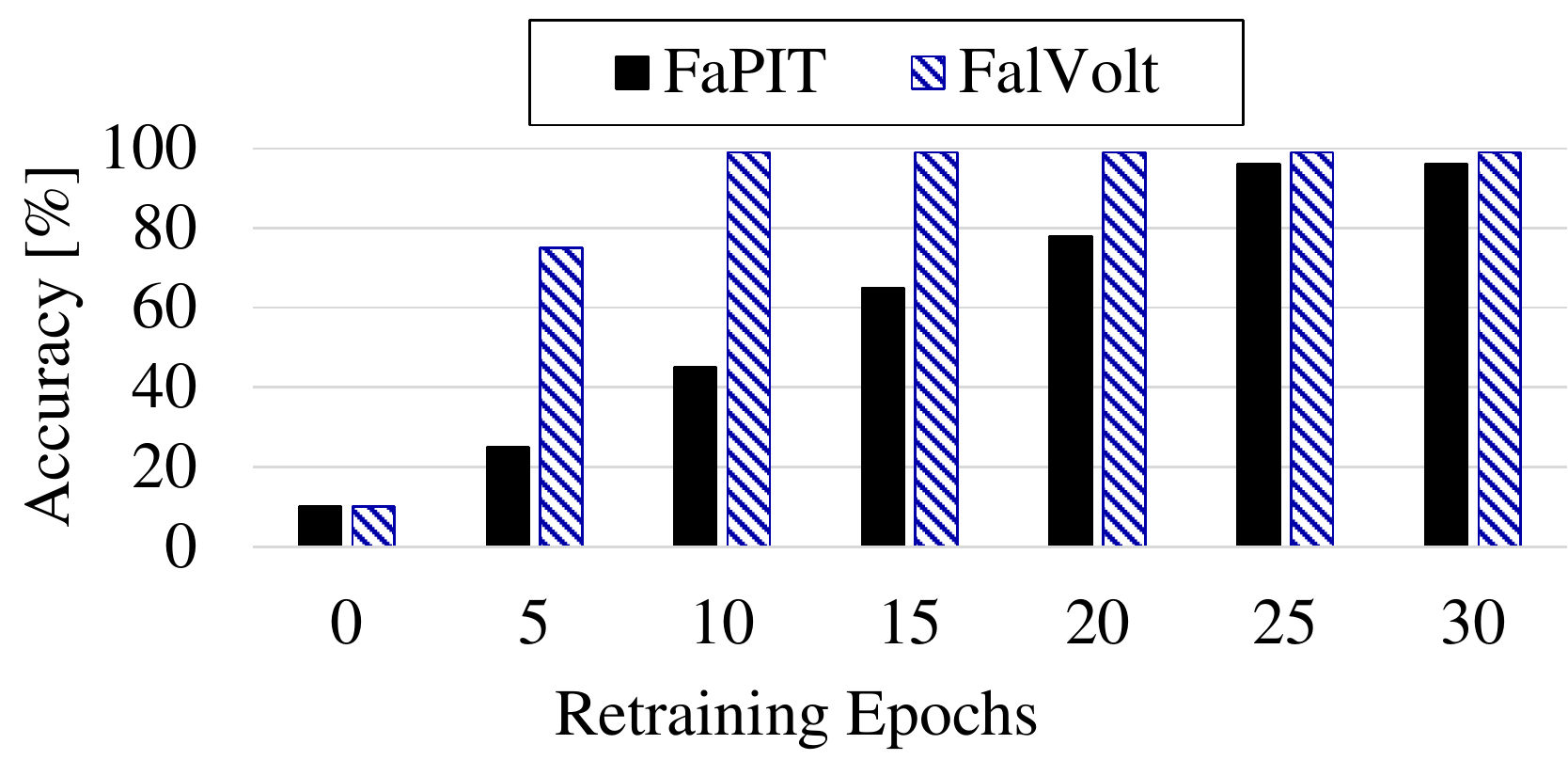}
         \caption{N-MNIST~\cite{orchard2015converting} classification}
         \label{subfig:mitepoch22}
     \end{subfigure}     
     \hfill
     \begin{subfigure}[b]{0.33\textwidth}
         \centering
         \includegraphics[width=1\textwidth]{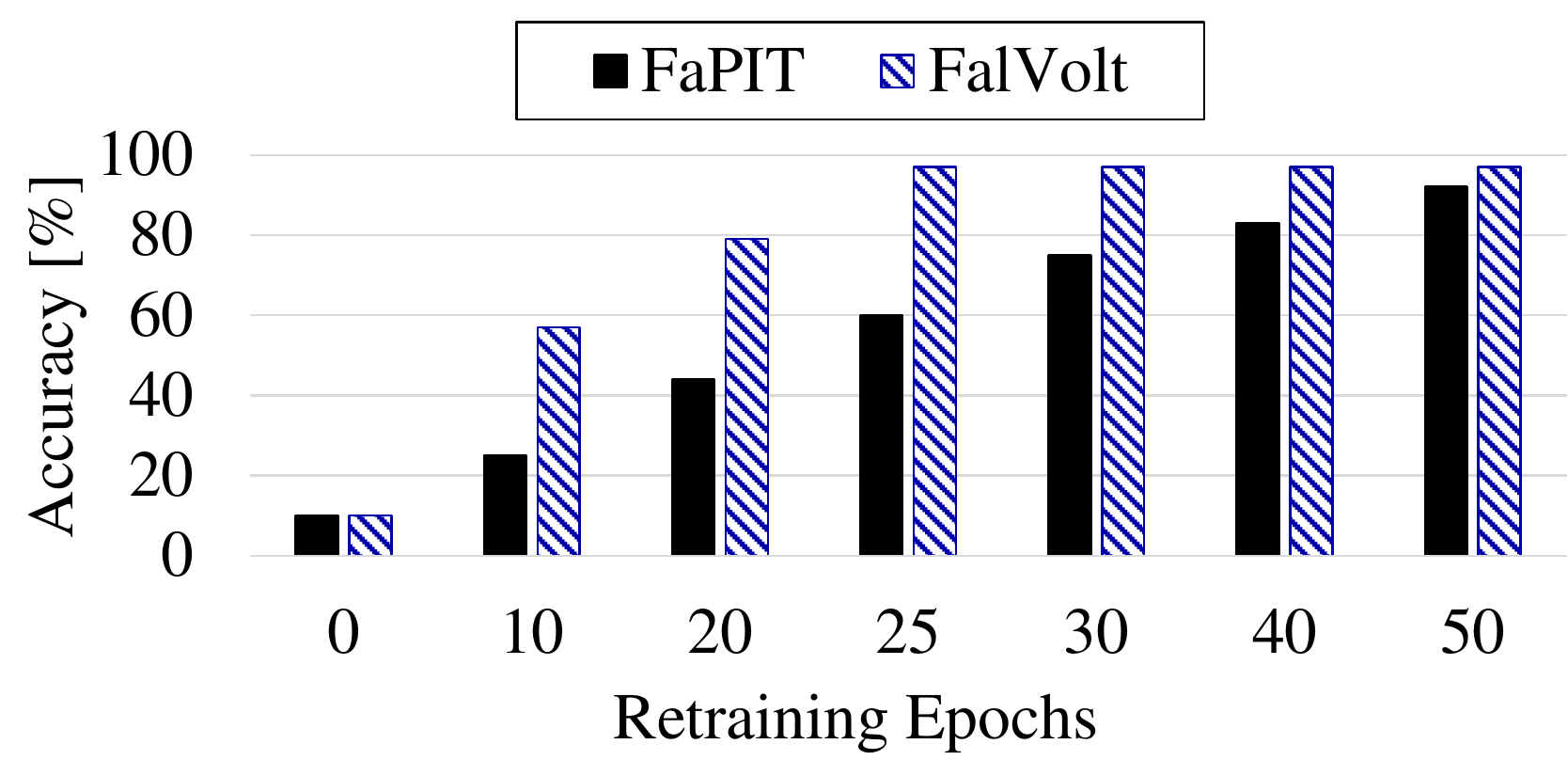}
         \caption{DVS128 Gesture~\cite{amir2017low} classification}
         \label{subfig:mitepoch33}
     \end{subfigure}
        \vspace{-0.05in}
        \caption{Performance of FaPIT and FalVolt over different epochs when 30\% the total PEs are faulty in a 256x256 systolic-array SNN accelerator (systolicSNN)}
        \label{fig:mitepoch}
        \end{adjustbox}
        \vspace{-0.2in}
\end{figure*} 

\uline{Classification accuracy vs. fault rates:} For the fault mitigation analysis, we inject the stuck-at faults using different fault maps in 10\%, 30\%, and 60\% PEs of a 256x256 systolicSNN and run paralleled re-training simulations. We employ the proposed FalVolt mitigation method using Algorithm 1 for 10\%, 30\%, and 60\% PEs in a 256x256 systolicSNN. Our analysis shows that optimizing threshold voltage for each hidden convolutional and fully connected layer helps in achieving baseline accuracy. Fig.~\ref{fig:mitconv} shows the optimized threshold voltage returned from the FalVolt mitigation method for each hidden layer to achieve the baseline accuracy for MNIST, NMNIST, and DVS128 Gesture datasets. For all these datasets, the optimized threshold voltage for the initial spiking-convolutional and spiking-fully connected layers is higher than other layers to ensure that the redundant spikes do not travel to the output layer. 

Fig.~\ref{fig:mitcomp} compares the FalVolt mitigation method with FaP and FaPIT. We observe that an increased fault rate causes a rapid accuracy loss in the FaP. FaPIT and FalVolt help in improving classification accuracy. However, only FalVolt achieves the baseline classification accuracy in the MNIST, N-MNIST, and DVS128 Gesture classification with even 60\% of the faulty PEs. This validates the applicability of FalVolt to both static and neuromorphic datasets.

\uline{Classification accuracy vs. number of epochs:} FalVolt increases the classification accuracy at the cost of additional retraining epochs to FaP; however, they are negligible compared to the lifetime of systolicSNNs. As shown in Fig.~\ref{fig:mitepoch}, FaPVolt is 2x faster than FaPIT. For example, the classification accuracy of MNIST is as high as 80\% with FaPIT using 20 epochs and converges with baseline accuracy around 25 epochs. However, the same dataset achieves the baseline accuracy with FalVolt in 10 epochs, as shown in Fig.~\ref{subfig:mitepoch11}. Likewise, FalVolt achieves the baseline accuracy of NMNIST classification 2x less number of epochs when compared to FaPIT as shown in Fig.~\ref{subfig:mitepoch22}. Moreover, the classification accuracy of DVS128 Gesture is as high as 83\% with FaPIT using 40 epochs and converges with baseline accuracy around 50 epochs as shown in Fig.~\ref{subfig:mitepoch33}. However, the same dataset achieves 97\% accuracy with FalVolt around 25 epochs. Since a small change in the baseline accuracy may cause catastrophic issues in safety-critical applications; therefore, the epochs for initial training, FaPIT, and FalVolt algorithms are high to achieve the classification accuracy close to the baseline. Note, training the large-sized SNNs itself takes a long time (or a higher number of epochs).

\section{Conclusion}
\label{sec:conclusion}
This paper extensively analyzes the stuck-at fault vulnerabilities of systolicSNNs and proposes a novel fault mitigation technique `\textit{fault-aware retraining through threshold voltage optimization (FalVolt)}.' FalVolt uses an optimized threshold voltage and time steps different from initial training to achieve classification accuracy close to the baseline. To demonstrate the effectiveness of FalVolt, we classify the MNIST, N-MNIST, and DVS128 Gesture datasets on a 256x256 systolicSNN while injecting faults at different rates. Our results show that even 0.012\% faulty PEs in a systolicSNN leads to significant accuracy loss. However, FalVolt improves the performance of systolicSNNs by enabling them to operate at fault rates of up to 60\%, with a negligible drop in the classification accuracy (as low as 0.1\%). Furthermore, our results show that FalVolt is 2x faster when compared to state-of-the-art techniques, such as fault-aware pruning without threshold voltage optimization.  


\bibliographystyle{IEEEtran}
\bibliography{bib/conf}

\begin{thebibliography}{10}
\providecommand{\url}[1]{#1}
\csname url@samestyle\endcsname
\providecommand{\newblock}{\relax}
\providecommand{\bibinfo}[2]{#2}
\providecommand{\BIBentrySTDinterwordspacing}{\spaceskip=0pt\relax}
\providecommand{\BIBentryALTinterwordstretchfactor}{4}
\providecommand{\BIBentryALTinterwordspacing}{\spaceskip=\fontdimen2\font plus
\BIBentryALTinterwordstretchfactor\fontdimen3\font minus
  \fontdimen4\font\relax}
\providecommand{\BIBforeignlanguage}[2]{{%
\expandafter\ifx\csname l@#1\endcsname\relax
\typeout{** WARNING: IEEEtran.bst: No hyphenation pattern has been}%
\typeout{** loaded for the language `#1'. Using the pattern for}%
\typeout{** the default language instead.}%
\else
\language=\csname l@#1\endcsname
\fi
#2}}
\providecommand{\BIBdecl}{\relax}
\BIBdecl

\bibitem{painkras2013spinnaker}
E.~Painkras~et al., ``Spinnaker: A 1-w 18-core system-on-chip for
  massively-parallel neural network simulation,'' \emph{IEEE Journal of
  Solid-State Circuits}, vol.~48, no.~8, pp. 1943--1953, 2013.

\bibitem{akopyan2015truenorth}
F.~Akopyan~et al., ``Truenorth: Design and tool flow of a 65 mw 1 million
  neuron programmable neurosynaptic chip,'' \emph{IEEE transactions on
  computer-aided design of integrated circuits and systems}, vol.~34, no.~10,
  pp. 1537--1557, 2015.

\bibitem{lee2020reconfigurable}
J.~J. Lee~et al., ``Reconfigurable dataflow optimization for spatiotemporal
  spiking neural computation on systolic array accelerators,'' in
  \emph{\textbf{ICCD}}.\hskip 1em plus 0.5em minus 0.4em\relax IEEE, 2020, pp.
  57--64.

\bibitem{guo2019systolic}
S.~Guo~et al., ``A systolic snn inference accelerator and its co-optimized
  software framework,'' in \emph{Proceedings of the 2019 on Great Lakes
  Symposium on VLSI}, 2019, pp. 63--68.

\bibitem{chuang202090nm}
P.~Y. Chuang~et al., ``A 90nm 103.14 tops/w binary-weight spiking neural
  network cmos asic for real-time object classification,'' in
  \emph{\textbf{DAC}}.\hskip 1em plus 0.5em minus 0.4em\relax IEEE, 2020, pp.
  1--6.

\bibitem{tan2020power}
P.~Y. Tan~et al., ``A power-efficient binary-weight spiking neural network
  architecture for real-time object classification,'' \emph{arXiv preprint
  arXiv:2003.06310}, 2020.

\bibitem{leeparallel}
J.~J. Lee~et al., ``Parallel time batching: Systolic-array acceleration of
  sparse spiking neural computation,'' \emph{\textbf{HPCA}}, pp. 317--330,
  2022.

\bibitem{kung2019packing}
H.~T. Kung~et al., ``Packing sparse convolutional neural networks for efficient
  systolic array implementations: Column combining under joint optimization,''
  in \emph{\textbf{ASPLOS}}, 2019, pp. 821--834.

\bibitem{el2020securing}
R.~e.~a. El-Allami, ``Securing deep spiking neural networks against adversarial
  attacks through inherent structural parameters,'' \emph{arXiv preprint
  arXiv:2012.05321}, 2020.

\bibitem{lee2021systolic}
J.~J. Lee~et al., ``Systolic-array spiking neural accelerators with dynamic
  heterogeneous voltage regulation,'' in \emph{\textbf{IJCNN}}.\hskip 1em plus
  0.5em minus 0.4em\relax IEEE, 2021, pp. 1--7.

\bibitem{schuman2020resilience}
C.~D. Schuman~et al., ``Resilience and robustness of spiking neural networks
  for neuromorphic systems,'' in \emph{\textbf{IJCNN}}.\hskip 1em plus 0.5em
  minus 0.4em\relax IEEE, 2020, pp. 1--10.

\bibitem{vatajelu2019special}
E.~I. Vatajelu~et al., ``Special session: Reliability of hardware-implemented
  spiking neural networks (snn),'' in \emph{\textbf{VTS}}.\hskip 1em plus 0.5em
  minus 0.4em\relax IEEE, 2019, pp. 1--8.

\bibitem{guo2021neural}
W.~Guo~et al., ``Neural coding in spiking neural networks: A comparative study
  for robust neuromorphic systems,'' \emph{Frontiers in Neuroscience}, vol.~15,
  p. 212, 2021.

\bibitem{el2020spiking}
S.~A. El-Sayed, ``Spiking neuron hardware-level fault modeling,'' in
  \emph{\textbf{IOLTS}}.\hskip 1em plus 0.5em minus 0.4em\relax IEEE, 2020, pp.
  1--4.

\bibitem{spyrou2022reliability}
T.~Spyrou~et al., ``Reliability analysis of a spiking neural network hardware
  accelerator,'' in \emph{\textbf{DATE}}, 2022.

\bibitem{putra2021respawn}
R.~V.~W. Putra~et al., ``Respawn: Energy-efficient fault-tolerance for spiking
  neural networks considering unreliable memories,'' in
  \emph{\textbf{ICCAD}}.\hskip 1em plus 0.5em minus 0.4em\relax IEEE, 2021, pp.
  1--9.

\bibitem{vidya2022softsnn}
------, ``Softsnn: Low-cost fault tolerance for spiking neural network
  accelerators under soft errors,'' \emph{arXiv preprint arXiv:2203.05523},
  2022.

\bibitem{venceslai2020neuroattack}
V.~Venceslai~et al., ``Neuroattack: Undermining spiking neural networks
  security through externally triggered bit-flips,'' in
  \emph{\textbf{IJCNN}}.\hskip 1em plus 0.5em minus 0.4em\relax IEEE, 2020, pp.
  1--8.

\bibitem{rastogi2021self}
M.~Rastogi~et al., ``On the self-repair role of astrocytes in stdp enabled
  unsupervised snns,'' \emph{Frontiers in Neuroscience}, vol.~14, p. 1351,
  2021.

\bibitem{siddique2021exploring}
A.~Siddique~et al., ``Exploring fault-energy trade-offs in approximate dnn
  hardware accelerators,'' in \emph{\textbf{ISQED}}.\hskip 1em plus 0.5em minus
  0.4em\relax IEEE, 2021, pp. 343--348.

\bibitem{zhang2018analyzing}
J.~J. Zhang~et al., ``Analyzing and mitigating the impact of permanent faults
  on a systolic array based neural network accelerator,'' in
  \emph{\textbf{VTS}}.\hskip 1em plus 0.5em minus 0.4em\relax IEEE, 2018, pp.
  1--6.

\bibitem{abdullah2020salvagednn}
M.~A. Hanif~et al., ``Salvagednn: salvaging deep neural network accelerators
  with permanent faults through saliency-driven fault-aware mapping,''
  \emph{Philosophical Transactions of the Royal Society A}, vol. 378, no. 2164,
  p. 20190164, 2020.

\bibitem{kundu2020high}
S.~Kundu~et al., ``High-level modeling of manufacturing faults in deep neural
  network accelerators,'' in \emph{\textbf{IOLTS}}.\hskip 1em plus 0.5em minus
  0.4em\relax IEEE, 2020, pp. 1--4.

\bibitem{cohen2017emnist}
Y.~LeCun~et al., ``Mnist handwritten digit database,'' \emph{ATT Labs [Online].
  Available: http://yann.lecun.com/exdb/mnist}, vol.~2, 2010.

\bibitem{orchard2015converting}
G.~Orchard, A.~Jayawant, G.~K. Cohen, and N.~Thakor, ``Converting static image
  datasets to spiking neuromorphic datasets using saccades,'' \emph{Frontiers
  in neuroscience}, vol.~9, p. 437, 2015.

\bibitem{amir2017low}
A.~Amir~et al., ``A low power, fully event-based gesture recognition system,''
  in \emph{\textbf{CVPR}}, 2017, pp. 7243--7252.

\bibitem{fang2021incorporating}
W.~Fang~et al., ``Incorporating learnable membrane time constant to enhance
  learning of spiking neural networks,'' in \emph{\textbf{ICCV}}, 2021, pp.
  2661--2671.

\bibitem{wang2020sies}
S.~Q. Wang~et al., ``Sies: A novel implementation of spiking convolutional
  neural network inference engine on field-programmable gate array,''
  \emph{Journal of Computer Science and Technology}, vol.~35, no.~2, pp.
  475--489, 2020.

\bibitem{morris2022hyperspike}
J.~Morris~et al., ``Hyperspike: hyperdimensional computing for more efficient
  and robust spiking neural networks,'' in \emph{\textbf{DATE}}.\hskip 1em plus
  0.5em minus 0.4em\relax IEEE, 2022, pp. 664--669.

\bibitem{lee2020enabling}
C.~Lee, S.~S. Sarwar, P.~Panda, G.~Srinivasan, and K.~Roy, ``Enabling
  spike-based backpropagation for training deep neural network architectures,''
  \emph{Frontiers in neuroscience}, p. 119, 2020.

\bibitem{hanif2020dependable}
M.~A. Hanif~et al., ``Dependable deep learning: Towards cost-efficient
  resilience of deep neural network accelerators against soft errors and
  permanent faults,'' in \emph{\textbf{IOLTS}}.\hskip 1em plus 0.5em minus
  0.4em\relax IEEE, 2020, pp. 1--4.

\end{thebibliography}

\end{document}